\documentclass[letterpaper, 10 pt, conference]{ieeeconf}  

\IEEEoverridecommandlockouts                              

\usepackage{hyperref}
\usepackage{graphics} 
\usepackage{epsfig} 
\usepackage{amsmath} 
\usepackage{amssymb}  
\usepackage{booktabs}
\usepackage{multirow}
\usepackage{color}
\newcommand{\I}{\textbf{\text{I}}}
\newcommand{\G}{\textbf{\text{G}}}
\newcommand{\st}{\textbf{\text{s}}}
\newcommand{\at}{\boldsymbol{a}}
\newcommand{\pt}{\boldsymbol{\theta}}
\newcommand{\ot}{\textbf{\text{o}}}

\title{\LARGE \bf
Learning Language-Conditioned Deformable Object Manipulation with Graph Dynamics
}

\author{Yuhong Deng$^{1,\dagger}$, Kai Mo$^{2,\dagger}$, Chongkun Xia$^{2}$ and Xueqian Wang$^{2,*}$ 
\thanks{$^\dagger$ indicates the authors with equal contributions.}
\thanks{ $^{1}$ National University of Singapore, Singapore}
\thanks{ $^{2}$  Tsinghua  Shenzhen  International Graduate School, Shenzhen, China }
\thanks{$^*$Corresponding author}%
}

\begin{document}

\maketitle
\thispagestyle{empty}
\pagestyle{empty}

\begin{abstract}
Multi-task learning of deformable object manipulation is a challenging problem in robot manipulation. Most previous works address this problem in a goal-conditioned way and adapt goal images to specify different tasks, which limits the multi-task learning performance and can not generalize to new tasks. Thus, we adapt language instruction to specify deformable object manipulation tasks and propose a learning framework. We first design a unified Transformer-based architecture to understand multi-modal data and output picking and placing action. Besides, we have introduced the visible connectivity graph to tackle nonlinear dynamics and complex configuration of the deformable object. Both simulated and real experiments have demonstrated that the proposed method is effective and can generalize to unseen instructions and tasks. Compared with the state-of-the-art method, our method achieves higher success rates (87.2\% on average) and has a 75.6\% shorter inference time. We also demonstrate that our method performs well in real-world experiments. Supplementary videos can be found at \url{https://sites.google.com/view/language-deformable}.

\end{abstract}

\section{INTRODUCTION}
\label{sec:intro}
Recently, vision-based deformable object manipulation has been widely investigated. The robot is supposed to infer a sequence of manipulation actions from visual observations to manipulate a deformable object into a prescribed goal configuration~\cite{pointcloud_df,rgb_df}. Early works focus on learning task-specific manipulation skills. A generic robot that can complete different deformable object manipulation tasks is still an open challenge. Toward this goal, some researchers have made progress in using goal images to specify different tasks and achieve multi-task learning~\cite{goal_image_1,goal_image_2}. However, goal images often over-defined tasks by information not related to the task requirement, such as position, color, and image style. Besides, the model can not deal with unseen goal images from different domains and generalize to new tasks.\par


\begin{figure}[t]
    \centering
    \includegraphics[width=0.485\textwidth]{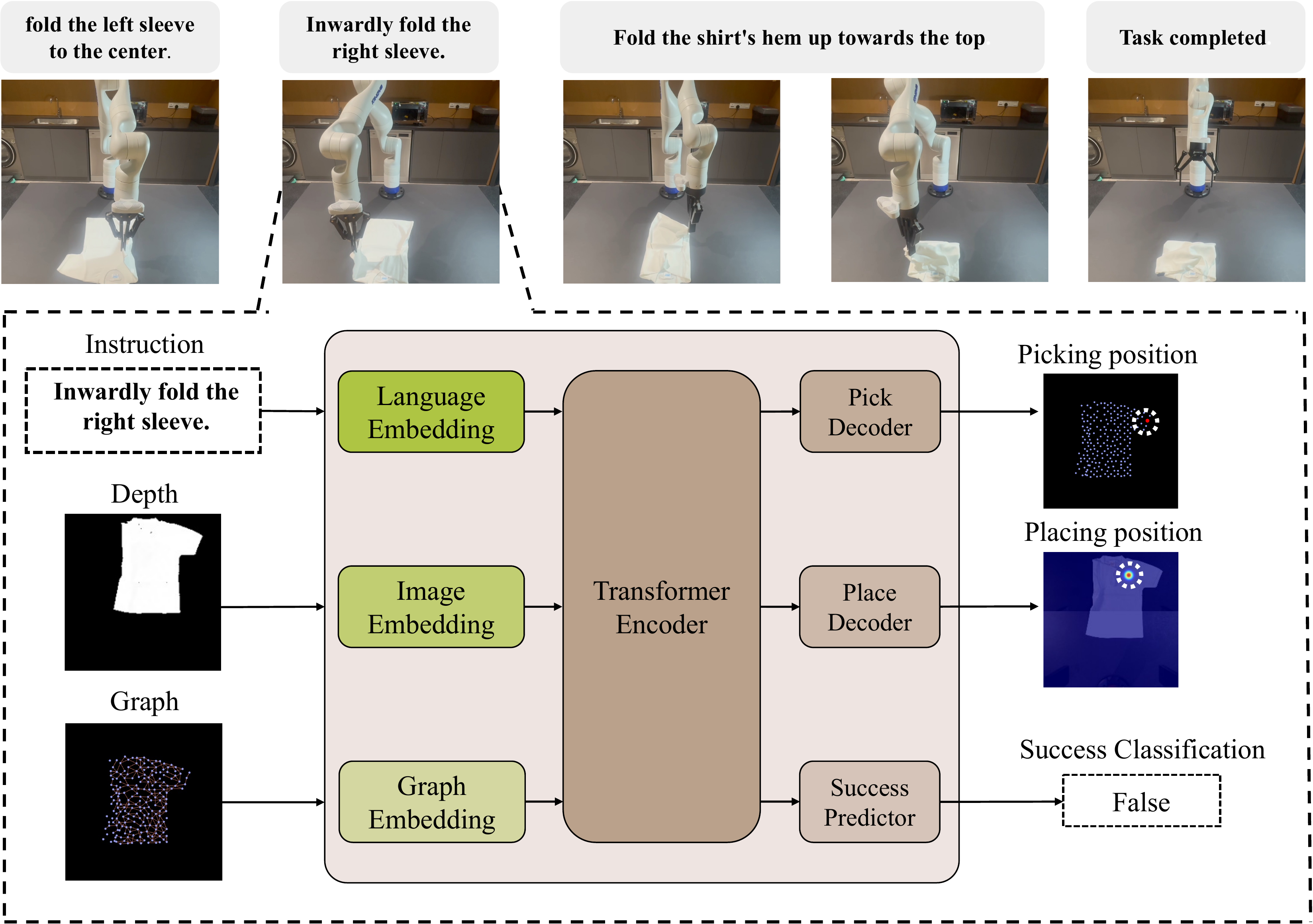}
    \caption{\textbf{Overview}. We design a unified Transformer-based model and introduce graph representation to solve language-conditioned deformable object manipulation tasks. Our model performs well on deformable object manipulation tasks.}
    \label{fig:overview}
    \vspace{-0.5cm}
\end{figure}

Compared with goal image, language instruction only provide necessary information related to task requirement and can define tasks flexibly. The prior knowledge of the pre-trained language model can also help the robot deal with unseen tasks. Thus, language-conditioned manipulation has been a new trend to pave the way for the generic robot that can complete various manipulation tasks. However, previous language-conditioned manipulation learning methods are limited to templated manipulations like picking, placing, and moving rigid objects~\cite{cliport,saycan}. Language-conditioned deformable object manipulation tasks have not been solved for two reasons. Deformable object manipulation tasks are often sequential multi-step tasks, which pose higher requirements for inference and decision-making at the intersection of vision, language, and action. In addition, the manipulation task is more difficult because the nonlinear dynamics~\cite{def_dynamic} and complex configurations~\cite{def_config} of deformable objects.\par

We design a unified Transformer-based model architecture to tackle the challenge of inference and decision-making on multi-modal data (Fig.~\ref{fig:overview}). We first implement a CLIP~\cite{clip} model pre-trained on millions of image-caption pairs to provide language embeddings, which can provide a powerful prior for grounding language in visual scenes. As for the image, we borrow the method in ViT~\cite{vit} (Vision Transformer) and get the visual embeddings by linear projection. To deal with deformable dynamics and configurations, we establish a visible connectivity graph~\cite{connect_graph} and get the graph embeddings by linearly projecting representation vectors of graph nodes. The visible connectivity graph can represent the deformable object's spatial structure and overcome the challenges of partial observability and self-occlusions. After obtaining the language, image, and graph embeddings, all embeddings will pass through the type embedding layer~\cite{type_embedding}  to achieve multi-modal data fusion. Finally, a Transformer~\cite{transformer} encoder-decoder structure is used to generate picking and placing action possibility distribution, where positions with max possibility will be picking and placing positions.


Since there is no existing dataset for language-conditioned deformable object manipulation tasks, we build a dataset including various
language instructions and corresponding deformable object manipulation tasks in the simulation environment. We then conduct simulated experiments to evaluate our proposed method. The results demonstrate that our method outperforms the state-of-the-art framework~\cite{cliport} on multi-task learning of deformable object manipulation. Our model can also generalize to unseen instructions and tasks not provided in the training process. Real-world experiments also demonstrate that our model trained in simulation can be zero-shot transferred to the real world.
The contributions of this work are summarized as follows:
\begin{itemize}
	\item [1)]
	We propose a novel learning framework that extends instruction-following robots' application on sequential multi-step deformable object manipulation;
 	\item [2)]
	We propose a unified Transformer-based model architecture for language-conditioned deformable object manipulation policy learning;
	\item [3)] We conduct simulated and real-world experiments to demonstrate that the proposed framework outperforms the state-of-the-art method and can generalize to unseen instructions and tasks.
\end{itemize}
The rest of this paper is organized as follows. The related work is reviewed in Section~\ref{sec:related_work}. 
Section~\ref{sec:approach} presents details of the learning framework design. The experimental setup and results are provided in Section~\ref{sec:experi}. Section~\ref{sec:conclu} concludes this paper and discusses future works.


\section{RELATED WORK}
\label{sec:related_work}
\subsection{Learning for Deformable Object Manipulation}
Learning-based methods have been widely used to equip the robot with advanced deformable object manipulation abilities. There are two main learning-based methods: Model-based methods rely on a forward dynamics model, which can predict the configurations of deformable objects under a given action. The critical issue is to obtain accurate dynamics. Learning accurate deformable dynamics from interaction data between robots and objects has become a solution~\cite{data-driven_modeling,def_mpc}. However, the generalization of the learning model on different objects and tasks is still an open challenge. Policy-based methods learn manipulation policy directly from observation without a forward dynamics model. The robot learns manipulation policy from expert demonstrations~\cite{def_il} or exploratory robot interactions~\cite{def_rl}. However, most previous policy-based methods are limited to task-specific policy, which is inefficient in real-world applications~\cite{rl_condi}. Some approaches learn a goal-conditioned policy to provide a general framework for deformable object manipulation, where the goal image specifies the task~\cite{goal_image_1,goal_image_2}. Unlike previous work, we adapt language instructions to specify different deformable object manipulation tasks and achieve a general learning framework.




\subsection{Language-conditioned Robotic Manipulation policy}
Language-conditioned robotic manipulation has received growing attention recently, where the manipulation task is specified by language instruction~\cite{stengel2022guiding,shridhar2022perceiver}. Language instruction specification is much easier to obtain than goal image specification and can specify more diversified tasks feasibly and effectively. Despite the abundant benefits of commanding robots with natural language, such agents require deep integration of multiple data modalities (language, vision, action). Language-Conditioned Imitation Learning addresses this problem by mapping actions directly from vision and language understanding in an end-to-end fashion~\cite{cliport}. However, collecting demonstrations paired with language annotations in the real robot is costly and time-consuming. To tackle this data-collecting problem, Nair et al.~\cite{nair2022learning} adopt crowd-sourced annotation to obtain sufficient datasets. Although some works improve the model's performance in more diverse languages~\cite{guhur2022instruction,mees2022calvin}, previous language-conditioned robotic manipulation approaches are limited in rigid objects. The robot only needs to manipulate a rigid object with low-level skills such as grasping and pushing. Simplifying objects and actions will limit the application scenarios. Thus, we propose a solution framework that can be applied to complex deformable object manipulation tasks.\par

\begin{figure*}[t]
    \centering
    \includegraphics[width=0.9\textwidth]{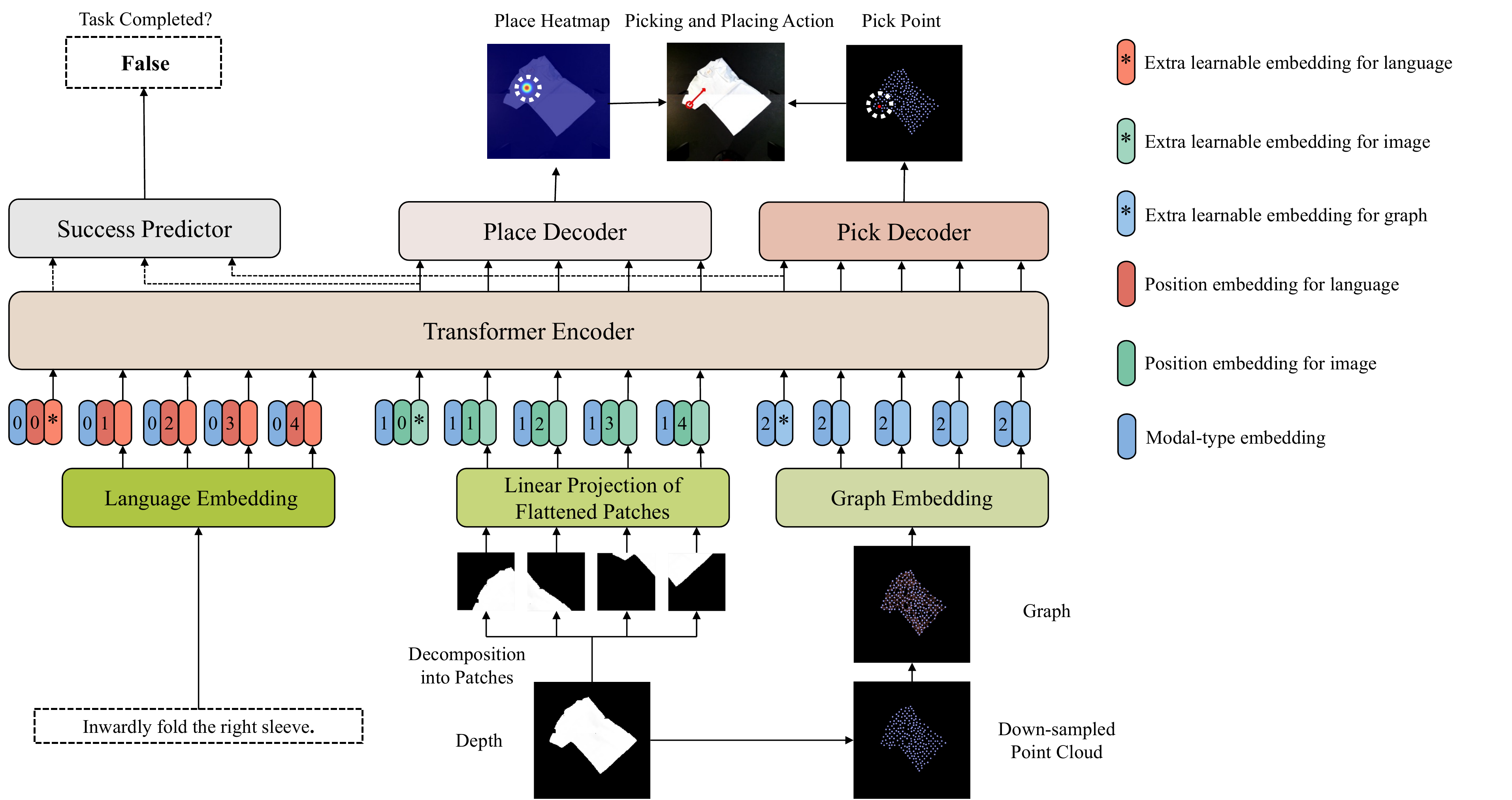}
    \caption{\textbf{Method overview.} We design a unified Transformer-based model architecture to understand the multi-modal data and output picking and placing action with task completion prediction. We introduce a visible connectivity graph to tackle deformable objects' complex configurations and dynamics.}
    \label{fig:method}
    \vspace{-0.5cm}
\end{figure*}


\section{Methods} 
\label{sec:approach}
\subsection{Problem Formulation}
\label{sec:problem}
Our goal is to learn a language-conditioned deformable object manipulation policy $\pi_{\pt} $ parameterized by $\pt$, which can generate a sequence of manipulation action $\{\at_{t} \} (t=0, 1,2,...T)$ in a closed-loop manner from a  language instruction $\st$ and the current visual observation $\ot_t$:
\begin{equation}
\label{eq:problem}
	\begin{split}
		&\at_{t} \leftarrow \pi_{\pt}(\ot_t, \st) \\
		\text{and}\quad	 &\ot_{t+1} \leftarrow\mathcal{T}(\ot_t, \at_{t})
	\end{split}
\end{equation}  

where $\mathcal{T}$ denotes the state transition. Since the deformable object has a complex configuration, the visual observation $\ot$ is composed of two parts in our solution framework: the top-down depth image $\I$ and the representation graph $\G$:
\begin{equation}
\label{eq:policy}
\at_{t} \leftarrow \pi_{\pt}(\I_t, \G_t, \st) 
\end{equation}  
The action space is defined as picking and placing action:
\begin{equation}\label{eq:action}
	\at_t = \{\at_{t}^{\text{pick}}, \at_{t}^{\text{place}}\}
\end{equation}
where $\at_{t}^{\text{pick}}$ and $\at_{t}^{\text{place}}$denote the robot end-effector's picking and placing poses. \par
We formulate the problem as a supervised learning problem. The policy $\pi_{\pt}$ is learned from expert demonstrations.

\subsection{Model Architecture}
We design a unified transformer-based model architecture (Fig.~\ref{fig:method}) for learning language-conditioned deformable object manipulation policy. We first leverage different modules to embed these multi-modal data. Then an encoder-decoder structure is used to generate picking and placing actions from multi-modal embeddings.\par
\textbf{Language Embedding:} 
We employ the language encoder in the CLIP model~\cite{clip} to embed the language instruction $\st$, which can provide a powerful prior for grounding language in visual scenes. Like previous works~\cite{vit,vilt}, we prepend an extra learnable embedding for aggregating the language representation and add learnable position embeddings to retain positional information. Mathematically, the language embedding $z_{s}$ is computed as follows:
\begin{equation}
    \label{eq:language embedding}
    z_{s} = [S_{head}; {\rm CLIP}(\st)] + S_{pos}
\end{equation}
where $\rm CLIP()$ denotes the CLIP's language encoder, $S_{head}$ denotes the extra learnable embedding, $S_{pos}$ denotes the position embeddings.

\textbf{Image Embedding:} Inspired by the previous work~\cite{foldsformer}, we use depth images rather than RGB images to make our framework transferable to the real world without fine-tune or domain randomization. We first decompose the depth image $\I \in \mathbb{R}^{H \times W \times 1}$ into $N$ non-overlapping patches. We then flatten these 2D patches into 1D vectors $x_{I} \in \mathbb{R}^{H \times W / N}$. We linearly map these vectors to align the CLIP output's dimension. We prepend an extra learnable embedding to aggregate the image representation and add learnable position embeddings for retaining positional information. Mathematically, The image embedding $z_{I}$ is computed as follows:
\begin{equation}
    z_I = [I_{head}; W_Ix_{I}^{1}; W_Ix_{I}^{2}; \cdots,W_Ix_{I}^{N}] + I_{pos}
\end{equation}
where $W_I$ denotes a learnable matrix, $I_{head}$ denotes the extra learnable embedding, $I_{pos}$ denotes the position embeddings.


\textbf{Graph Embedding:} We introduce a visible connectivity graph$\langle V,E \rangle$ to pose the inductive bias of the deformable objects' physics to the framework~\cite{connect_graph}. The nodes $V$ represent the particles that compose the deformable object. Specifically, the down-sampled point cloud $\textbf{\text{P}} = \{v_i\}_{i=1,\cdots,K}$ of the deformable obejct constructs the nodes $V$. There are two types of edges $E$, nearby edges $E^C$ model the collision between nodes and mesh edges $E^M$ model the deformable object's spatial structure. We construct nearby edges $E^C$ based on the euclidean distance of every two nodes firstly:
\begin{equation}
    E^C = \{e_{ij} | \Vert v_i - v_j \Vert_2 < R\}
    \label{eq:nearby_edges}
\end{equation}
where $R$ is a distance threshold, and $v_i$, $v_j$ are the positions of the nodes $i$, $j$. \par

We then use the pre-trained edge GNN \cite{connect_graph} to predict the mesh edges $E^M$ from the graph $\langle \textbf{\text{P}},E^C \rangle$. We also prepend an extra learnable embedding that can aggregate the graph representation. We do not add position embeddings because Transformer without position embeddings is permutation-invariant and a natural fit for graphs~\cite{no_position_emb}.
Mathematically, The graph embedding $z_G$ is computed as follows:
\begin{equation}
    z_G = [G_{head}; {G_{edge}}(\langle \textbf{\text{P}},E^C \rangle)]
\end{equation}
where $G_{head}$ denotes the extra learnable embedding, $G_{edge}()$ denotes the pre-trained edge GNN.

\begin{figure*}[t]
    \centering
    \includegraphics[width=0.9\textwidth]{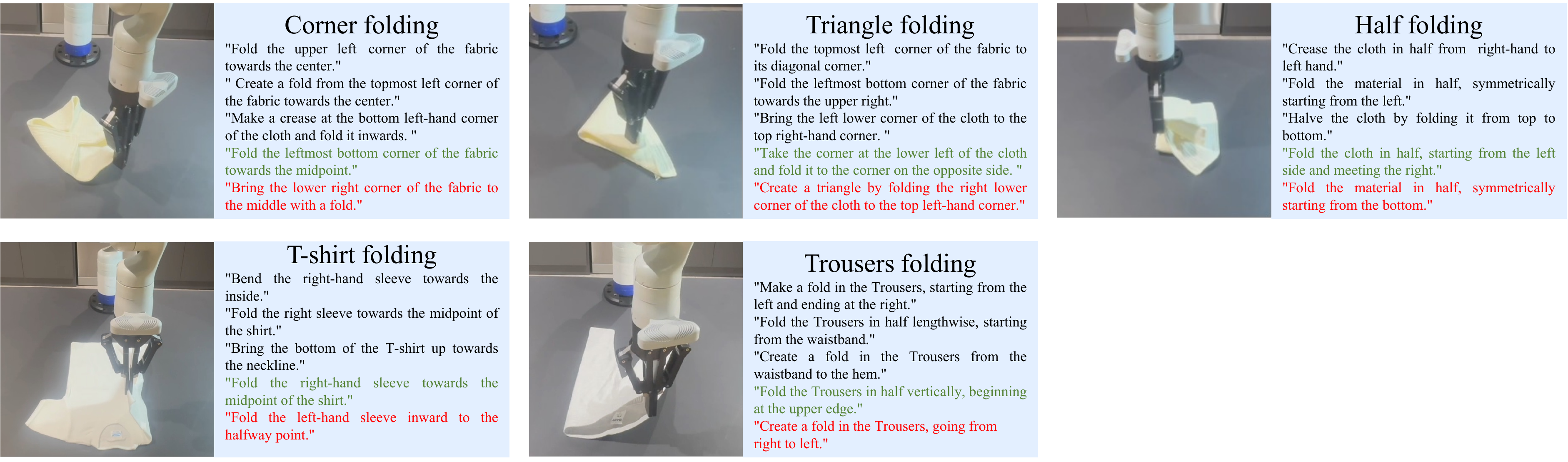}
    \caption{\textbf{Some examples of language-conditioned deformable object manipulation Tasks.} Seen instructions, unseen instructions, unseen tasks are marked in black, grey and red, respectively. }
    \label{fig:dataset}
    \vspace{-0.5cm}
\end{figure*}

\textbf{Encoder:}
After obtaining the multi-modal embeddings, we make an aggregation by concatenating these multi-modal embeddings and adding modal-type embeddings:
\begin{equation}
    z_0 = [z_s + S_{type}; z_{I}+I_{type}; z_G+G_{type}]
\end{equation}
where $S_{type}, I_{type}, G_{type}$ are modal-type embeddings of the language, the image and the graph, respectively.

We then input the vector $z_0$ to a Transformer encoder proposed in ViT~\cite{vit}, which has $L$ layers:
\begin{equation}
\begin{aligned}
    z^{\prime}_{l} &= {\rm MSA}({\rm LN}(z_{l-1})) + z_{l-1} \quad &l=1,\cdots,L\\
    z_l &= {\rm MLP}({\rm LN}(z^{\prime}_l)) + z^{\prime}_l \quad &l=1,\cdots,L
\end{aligned}
\end{equation}
where $\rm MSA()$ denotes multiheaded self-attention, $\rm LN()$ denotes LayerNorm, and $\rm MLP()$ denotes an MLP layer.

\textbf{Pick Decoder and Place Decoder:}
We then design a pick decoder and a place decoder to predict picking and placing position from the output embeddings of the encoder layer. The pick decoder is an MLP containing one layer, while the place decoder consists of convolutional and upsampling layers alternately. The pick decoder takes the graph's node embeddings as input and outputs the probability distribution $Q_{\rm pick} \in \mathbb{R}^K$, which represent the probability of each point in the down-sampled point cloud $\textbf{\text{P}}$ being a picking position. The place encoder takes the image's patch embeddings as input and outputs a pixel-wise heatmap $Q_{\rm place} \in \mathbb{R}^{H \times W}$, which represent the probability of each pixel in the image being a placing position. The optimal picking action$\boldsymbol{a}_{\rm pick}$ and the optimal placing action $\boldsymbol{a}_{\rm place}$ are computed as follows:

\begin{equation}
\begin{aligned}
    \boldsymbol{a}_{\rm pick} &= {\rm argmax}_{\boldsymbol{a}}Q_{\rm pick}(\boldsymbol{a}) \\
    \boldsymbol{a}_{\rm place} &= {\rm argmax}_{\boldsymbol{a}}Q_{\rm place}(\boldsymbol{a})
\end{aligned}
\end{equation}

\textbf{Success Classifier:}
Most previous language-conditioned manipulation learning algorithm can not estimate whether a task has been completed~\cite{cliport,shridhar2022perceiver,guhur2022instruction}, which is not practical in real-world employment. Thus, we design a successful classifier to estimate task completion, making our framework more autonomous. The success classifier is an MLP that contains two layers, which can perform binary classification to indicate task completion. The success classifier's input is the concatenation of the output embeddings corresponding to three extra embeddings $S_{head}$, $I_{head}$ and $G_{head}$. 

\subsection{Implementation Details:}

We use behavioral cloning to train a multi-task model. The training data is collected in the SoftGym suite~\cite{softgym}. The 3D models of deformable objects are sampled from CLOTH3D dataset~\cite{cloth3d}. In SoftGym, the deformable objects are modeled as particles whose ground truth positions and velocities can be accessed. Thus, we can easily collect expert demonstrations using a oracle demonstrator. The object's initial configurations (size, pose, etc.) are randomized during data collection. Each expert demonstration $\zeta$ is composed of the language instruction $\st$, the observation-action pair $(\ot_t,\at_t)$ with $t = 1, 2,\cdots, T$:\par
\begin{equation}
    \zeta = \{l,(\ot_1,\at_1),(\ot_2,\at_2),\cdots,(\ot_T,\at_T)\}
\end{equation}
For model training, we train the success classifier and other modules separately. We first freeze the CLIP encoder and the edge GNN and then train the modules except for the classifier with the binary cross-entropy (BCE) loss between the predicted $Q_{\rm pick}, Q_{\rm place}$ and the ground truth $Q_{\rm pick}^{gt}, Q_{\rm place}^{gt}$:
\begin{equation}
 \rm   \mathcal{L}_{action} = BCE(\textit{Q}_{pick},\textit{Q}_{pick}^{gt}) + BCE(\textit{Q}_{place},\textit{Q}_{place}^{gt})
\end{equation}
Finally, We freeze the trained modules and train the success classifier with the BCE loss.\par

\begin{table*}[t]
\vspace{0.2cm}
\centering
\small
\caption{\textbf{Simulation Experiment Results.} The average success rates (\%) on testing tasks (seen instructions, unseen instructions and unseen tasks). Models are trained with 100 and 1000 demonstrations per task. The best performance is in bold.}
\begin{tabular}{@{}lcccccccccc@{}}
\toprule
\multirow{2}*{Method} 
& \multicolumn{2}{c}{\begin{tabular}[c]{@{}c@{}}corner folding\\(seen instructions)  \end{tabular}} 
& \multicolumn{2}{c}{\begin{tabular}[c]{@{}c@{}} triangle folding\\(seen instructions) \end{tabular}} 
& \multicolumn{2}{c}{\begin{tabular}[c]{@{}c@{}}half folding\\(seen instructions) \end{tabular}} 
& \multicolumn{2}{c}{\begin{tabular}[c]{@{}c@{}}T-shirt folding\\(seen instructions) \end{tabular}}
& \multicolumn{2}{c}{\begin{tabular}[c]{@{}c@{}}Trousers folding\\(seen instructions) \end{tabular}} \\
\cmidrule(lr){2-3} \cmidrule(lr){4-5} \cmidrule(lr){6-7} \cmidrule(lr){8-9} \cmidrule(lr){10-11}
& 100 & 1000 & 100 & 1000 & 100 & 1000 & 100 & 1000 & 100 & 1000 \\
\midrule
Foldsformer~\cite{foldsformer} & 80.0 & 90.0 & 52.0 & 68.0 & 16.0 & 26.0 & 18.0 & 24.0 & 8.0 & 20.0 \\
CLIPORT~\cite{cliport} & 78.0 & 86.0 & \textbf{76.0} & 78.0 & 50.0 & 56.0 & 54.0 & 74.0 & 38.0 & 46.0 \\
ours (w/o graph) & 100,0 & 100.0 & 66.0 & 84.0 & 44.0 & 56.0 & 78.0 & 78.0 & 64.0 & \textbf{88.0} \\
ours (full method) & \textbf{100.0} & \textbf{100.0} & 72.0 & \textbf{92.0} & \textbf{52.0} & \textbf{74.0} & \textbf{86.0} & \textbf{84.0} & \textbf{74.0} & 86.0 \\

\midrule
\multirow{2}*{Method} 
& \multicolumn{2}{c}{\begin{tabular}[c]{@{}c@{}}corner folding\\(unseen instructions) \end{tabular}} 
& \multicolumn{2}{c}{\begin{tabular}[c]{@{}c@{}} triangle folding\\(unseen instructions) \end{tabular}} 
& \multicolumn{2}{c}{\begin{tabular}[c]{@{}c@{}}half folding\\(unseen instructions) \end{tabular}} 
& \multicolumn{2}{c}{\begin{tabular}[c]{@{}c@{}}T-shirt folding\\(unseen instructions) \end{tabular}}
& \multicolumn{2}{c}{\begin{tabular}[c]{@{}c@{}}Trousers folding\\(unseen instructions) \end{tabular}} \\
\cmidrule(lr){2-3} \cmidrule(lr){4-5} \cmidrule(lr){6-7} \cmidrule(lr){8-9} \cmidrule(lr){10-11}
& 100 & 1000 & 100 & 1000 & 100 & 1000 & 100 & 1000 & 100 & 1000 \\
\midrule
CLIPORT~\cite{cliport} & 80.0 & 90.0 & 76.0 & 74.0 & 38.0 & 50.0 & 56.0 & 70.0 & 32.0 & 38.0 \\
ours (w/o graph) & \textbf{100.0} & 100.0 & 72.0 & 78.0 & 36.0 & 60.0 & \textbf{86.0} & 80.0 & 54.0 & \textbf{84.0} \\
ours (full method) & 96.0 & \textbf{100.0} & \textbf{80.0} & \textbf{92.0} & \textbf{40.0} & \textbf{68.0} & 80.0 & \textbf{80.0} & \textbf{74.0} & \textbf{92.0} \\

\midrule
\multirow{2}*{Method} 
& \multicolumn{2}{c}{\begin{tabular}[c]{@{}c@{}}corner folding\\(unseen tasks) \end{tabular}} 
& \multicolumn{2}{c}{\begin{tabular}[c]{@{}c@{}} triangle folding\\(unseen tasks) \end{tabular}} 
& \multicolumn{2}{c}{\begin{tabular}[c]{@{}c@{}}half folding\\(unseen tasks) \end{tabular}} 
& \multicolumn{2}{c}{\begin{tabular}[c]{@{}c@{}}T-shirt folding\\(unseen tasks) \end{tabular}}
& \multicolumn{2}{c}{\begin{tabular}[c]{@{}c@{}}Trousers folding\\(unseen tasks) \end{tabular}} \\
\cmidrule(lr){2-3} \cmidrule(lr){4-5} \cmidrule(lr){6-7} \cmidrule(lr){8-9} \cmidrule(lr){10-11}
& 100 & 1000 & 100 & 1000 & 100 & 1000 & 100 & 1000 & 100 & 1000 \\
\midrule
CLIPORT~\cite{cliport} & 70.0 & 76.0 & 70.0 & 74.0 & 0.0 & 0.0 & 22.0 & 46.0 & 0.0 & 8.0 \\
ours (w/o graph) & \textbf{74.0} & \textbf{80.0} & 50.0 & 80.0 & 0.0 & \textbf{2.0} & 28.0 & 76.0 & \textbf{10.0} & 14.0 \\
ours (full method) & 36.0 & 78.0 & \textbf{76.0} & \textbf{88.0} & \textbf{0.0} & 0.0 & \textbf{42.0} & \textbf{80.0} & 0.0 & \textbf{32.0} \\

\bottomrule
\end{tabular}
\label{tab:exp_resu}
\vspace{-0.25cm}
\end{table*}

\section{Experiments}
\label{sec:experi}
This section presents simulated and real-world experiments to answer the following questions: 1) How well does our framework perform on multi-task learning of deformable object manipulation compared with the baseline methods? 2) What role does the visible connectivity graph play in our algorithm design? 3) How well does our framework generalize to unseen instructions and tasks? and 4) How well does our framework perform on real-world language-conditioned deformable object manipulation tasks?

\subsection{Simulation Experiments Setup}
All simulation experiments are conducted in the SoftGym suite~\cite{softgym}. The robot is provided with language instruction and is supposed to complete the instruction with only current visual observation inputs. We evaluate the performance of our framework with 5 types of language-conditioned manipulation tasks. Fig~\ref{fig:dataset} shows some examples, and TABLE~\ref{tab:testing_instruction}  shows the number of instructions associated with each type of task. There are various instructions and tasks in the simulation experiments. The testing instructions can be divided into three parts: seen instructions, unseen instructions, and unseen tasks. Seen instructions means instructions seen in the training process. The manipulation tasks that unseen instructions and seen instructions specify are the same. The difference is the language description. Unseen instructions can not be seen in the training process. Unseen tasks mean new tasks not seen in training. Unseen tasks require new manipulation skills that the model has not learned in training. For example, for corner folding, folding from the bottom left, top right, and top left can be seen in training; folding from the bottom right is an unseen task. For T-shirt folding, folding the right sleeve can be seen; folding the left sleeve is an unseen task.\par

\begin{table}[h]
\centering
\caption{\textbf{Testing instructions in simulation experiments.}}
\setlength{\tabcolsep}{2mm}{
\begin{tabular}{@{}lccc@{}}
\toprule
Task                          & seen instructions  & unseen instructions & unseen tasks \\ \midrule
corner folding    & 192 & 48 & 64  \\
triangle folding    & 192 & 48 & 64   \\
half folding    & 192 & 48 & 64  \\
T-shirt folding    & 96 & 24 & 48   \\
Trousers folding    & 96 & 24 & 48   \\
Total   & 768 & 192 & 288   \\
 \bottomrule
\end{tabular}
}
\label{tab:testing_instruction}
\vspace{-0.5cm}
\end{table}


\subsection{Simulation Experiment Results}
We compare the performance of our methods with two baseline methods: Foldsformer~\cite{foldsformer} is a state-of-the-art method that adapts a sequence of sub-goal images to specify the task. Foldsformer is provided with sub-goal images rather than language instruction in each task. CLIPORT~\cite{cliport} represents the typical network for language-conditioned manipulation learning. It relies on a two-stream architecture and uses pre-trained vision-language models for language-conditioned manipulation policies. Besides, we set an ablation study to evaluate the role of the graph in our framework. Ours (w/o graph) has the same backbone architecture as the proposed algorithm. The only difference is that we did not provide it with graph information. \par

We first evaluate the success rate on seen instructions, unseen instructions, and unseen tasks (each type of task has 50 task instances). We train all models with 100 and 1000 demonstrations separately. The success metric is the mean particle position error between the cloth states achieved by the policy and an oracle demonstrator. We define a task as a success if the mean particle position error is less than 0.0125m (the diameter of a particle in SoftGym). The results are shown in TABLE~\ref{tab:exp_resu}. \par
Overall, Foldsformer performs worst, illustrating the merits of language specification in multi-task learning of robot manipulation. Compared with a demonstration of an image sequence, natural language can provide sufficient cues of task requirements without over-defining the task by the object's texture, position, and size. Besides, a demonstration of an image sequence could only capture one instance of success~\cite{one_demo}.\par
Ours (w/o graph) generally outperforms CLIPORT, demonstrating that the proposed model architecture effectively learns language-conditioned manipulation policy. 
We use a unified Transformer-based model architecture to deal with multi-modal input and output picking and placing action. Such model architecture outperforms the state-of-the-art method.\par


\begin{table}[t]
\centering
\caption{\textbf{Model Capacity.} We compare model parameters, FLOPs, and inference time of different models.The best performance is in bold.}
\setlength{\tabcolsep}{2mm}{
\begin{tabular}{@{}lccc@{}}
\toprule
Method   & Params(M)  & FLOPS(G) & Inference Time (ms) \\ \midrule
Foldsformer~\cite{foldsformer}    & \textbf{11.6} & 24.58 & 19  \\
CLIPORT~\cite{cliport}   & 423.19 & 384.37 & 131   \\
ours (w/o graph)   & 51.03 & \textbf{19.81} & \textbf{15} \\
ours (full method)  & 53.55 & 34.38 & 32  \\
 \bottomrule
\end{tabular}
}
\label{tab:model_capacity}
\vspace{-0.5cm}
\end{table}

\begin{figure*}[t]
    \vspace{0.2cm}
    \centering
    \includegraphics[width=0.95\textwidth]{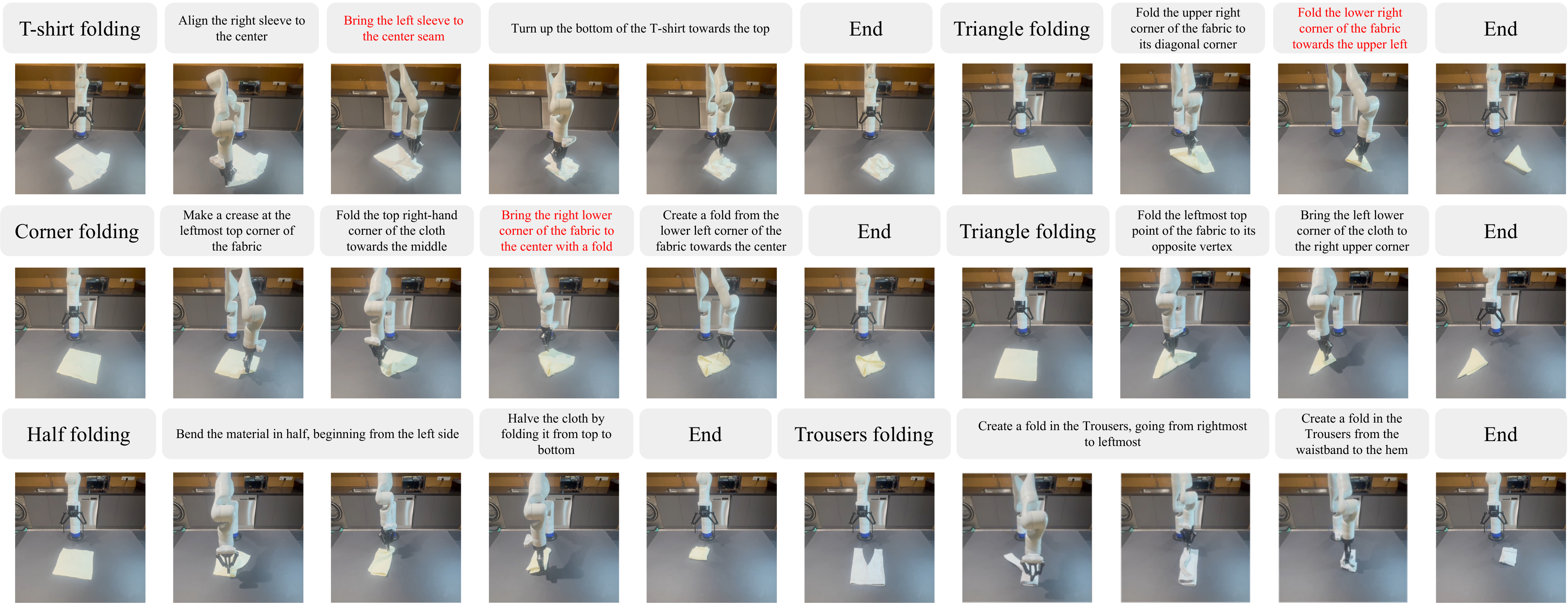}
    \caption{\textbf{Real World Experiments.} Our model performs well in language-conditioned deformable object manipulation tasks and can generalize to unseen tasks in the real world. Unseen tasks are marked in red.}
    \label{fig:real_world}
    \vspace{-0.5cm}
\end{figure*}

By introducing a visible connectivity graph, ours (full method) outperforms all baseline methods, especially on tasks with more steps and complex deformable objects (such as T-shirt and trousers folding). In these more complicated tasks, the manipulation process is more likely to cause irregular self-occlusion and partial observation. The visible connectivity graph can help the robot accurately capture deformable dynamics and configurations. \par
In addition, our model is also more data-efficient. In TABLE~\ref{tab:exp_resu}, the success rates of our model trained on 100 demonstrations are higher than those of baseline models trained on 1000 demonstrations on almost all involved tasks. \par
The experiment results also show that our model can be general to unseen instructions and even unseen tasks. Our model can understand unseen instructions with the prior knowledge of the pre-trained language model. More importantly, our model learns to ground the spatial displacement hidden in the instruction in the visual image instead of just memorizing some picking and placing positions, which makes the model general to unseen tasks such as folding from different directions and folding different parts. We also found that it is hard for the model to generalize to unseen tasks such as half folding, where the manipulation is more likely to lead to an arbitrary state. In these tasks, the model must see enough arbitrary visual features in the training.\par

Besides, our model is much lighter than the baseline method. We calculate different models' FLOPs (Floating Point Operations), model parameters, and inference times. The results are shown in TABLE.~\ref{tab:model_capacity}. Overall, Foldsformer is the lightest because it does not deal with multi-modal data. Ours (full method) and ours (w/o graph) are dramatically lighter in model FLOPs, parameters, and inference time than CLIPORT, which is attributed to the proposed unified transformer-based architecture. It can also be seen that improving model performance by introducing graph representation is feasible in terms of model efficiency. \par

\subsection{Real World Experiments}
As described previously, we use depth images rather than RGB images to make our framework able to be transferred to the real world directly. We evaluate the sim-to-real performance of our framework on a kinova robot with a standard two-finger Robotiq gripper. A Realsense RGB-D camera mounted on the robot's end-effector is used to capture visual observations. The deformable object (towel, T-shirt, or trousers) is placed on the platform in front of the robot. We evaluate the success rate of our model on 5 types of real-world language-conditioned manipulation tasks. We define a task as success according to the Mean Intersection Union (MIoU) between the deformable object masks achieved by our model and the human experts. If the MIou exceeds 0.9, the task is successful. Table.~\ref{tab:real world results} shows the results. Real experiments' success rate is close to simulation experiments' success rate.

\begin{table}[t]
\centering
\caption{\textbf{Real Experiment Results.}
The average success rate (\%) on testing tasks.
}
\setlength{\tabcolsep}{10mm}{
\begin{tabular}{@{}lc@{}}
\toprule
Task     & success rate(\%) \\ 
\midrule
corner folding    & 80.0  \\
triangle folding    & 70.0  \\
half folding    & 40.0 \\
T-shirt folding    & 60.0  \\
Trousers folding    & 40.0  \\  
 \bottomrule
\end{tabular}
}
\label{tab:real world results}
\vspace{-0.5cm}
\end{table}

Fig.~\ref{fig:real_world}  shows some examples in our real-world experiments.  Our model can complete both seen and unseen tasks in real-world experiments. A complete video recording of these experiments can be found on our project website.




\section{Conclusion}
\label{sec:conclu}
In this paper, we propose a novel framework for language-conditioned deformable manipulation policy learning. We design a unified Transformer-based model architecture to deal with multi-modal data and output precise picking and placing action. Besides, we construct a visible connective graph to represent the spatial structure of the deformable object. Extensive experiments have been conducted to verify the proposed model architecture's performance, proving that our framework can improve the multi-task learning performance on deformable object manipulation and generalize to unseen instructions and unseen tasks.
 Furthermore,  our framework performs well on real-world experiments. For future work, we will explore the method to deal with arbitrary configurations during manipulation and improve the proposed framework's performance on more complicated unseen tasks.
 


\bibliographystyle{ieeetr}
\bibliography{root.bbl}

\end{document}